\documentclass[10pt,twocolumn,letterpaper]{article}

\usepackage{iccv}
\usepackage{times}
\usepackage{epsfig}
\usepackage{graphicx}
\usepackage{amsmath}
\usepackage{amssymb}
\usepackage{booktabs}

\usepackage{algorithm}
\usepackage{algpseudocode}
\usepackage{threeparttable}
\usepackage{multirow}
\usepackage{booktabs}
\usepackage{caption}
\usepackage{mathrsfs}
\usepackage{bbding}
\usepackage{blindtext}

\usepackage{subcaption}
\usepackage[dvipsnames]{xcolor}
\usepackage[safe]{tipa} 
\usepackage{color, colortbl}
\newcommand{\tableCellHeight}{1}
\newcommand{\tabstyle}[1]{
	\setlength{\tabcolsep}{#1}
	\renewcommand{\arraystretch}{\tableCellHeight}
	\centering
	\small
}
\newcommand{\rotbox}[1]{\rotatebox{90}{#1}}

\definecolor{tabhighlight}{HTML}{e5e5e5}
\definecolor{citecolor}{HTML}{0071bc}

\usepackage[pagebackref=true,breaklinks=true,letterpaper=true,colorlinks,bookmarks=false]{hyperref}
 \iccvfinalcopy 

\def\iccvPaperID{2174} 

\ificcvfinal\fi

\begin{document}

\title{Gradient-Regulated Meta-Prompt Learning\\ for Generalizable Vision-Language Models}

\author{Juncheng Li$~\textsuperscript{\rm 1, 2}$\thanks{Equal Contribution.} \ \thanks{Work done when interning at Huawei Cloud.} \and Minghe Gao$~\textsuperscript{\rm 1}$\footnote[1]{}  \and  Longhui Wei$~\textsuperscript{\rm 2, 3}$ \and  Siliang Tang$~\textsuperscript{\rm 1}$   \and  Wenqiao Zhang$~\textsuperscript{\rm 4}$  \and   Mengze Li$~\textsuperscript{\rm 1}$ \and Wei Ji$~\textsuperscript{\rm 4}$  \and Qi Tian$~\textsuperscript{\rm 2}$ \and Tat-Seng Chua$~\textsuperscript{\rm 4}$ \and Yueting Zhuang$~\textsuperscript{\rm 1}$\thanks{Corresponding Author.} \\
	\small{$~\textsuperscript{\rm 1}$ Zhejiang University}, 
	\small{$~\textsuperscript{\rm 2}$ Huawei Cloud},
	\small{$~\textsuperscript{\rm 3}$ University of Science and Technology of China}
	\small{$~\textsuperscript{\rm 4}$ National university of Singapore}
	\\{\tt\small {\{junchengli, 22221320, siliang, mengzeli, yzhuang\}}@zju.edu.cn}
	\\{\tt\small {\{weilonghui1, tian.qi1\}}@huawei.com, {\{wenqiao, jiwei, dcscts\}}@nus.edu.sg}
}
\maketitle
\ificcvfinal\thispagestyle{empty}\fi

\begin{abstract}
	Prompt tuning, a recently emerging paradigm, enables the powerful vision-language pre-training models to adapt to downstream tasks in a parameter- and data- efficient way, by learning the ``soft prompts'' to condition frozen pre-training models. Though effective, it is particularly problematic in the few-shot scenario, where prompt tuning performance is sensitive to the initialization and requires a time-consuming process to find a good initialization, thus restricting the fast adaptation ability of the pre-training models. In addition, prompt tuning could undermine the generalizability of the pre-training models, because the learnable prompt tokens are easy to overfit to the limited training samples. To address these issues, we introduce a novel Gradient-RegulAted Meta-prompt learning~(GRAM) framework that jointly meta-learns an efficient soft prompt initialization for better adaptation and a lightweight gradient regulating function for strong cross-domain generalizability in a meta-learning paradigm using only the unlabeled image-text pre-training data. Rather than designing a specific prompt tuning method, our GRAM can be easily incorporated into various prompt tuning methods in a model-agnostic way, and comprehensive experiments  show that GRAM brings about consistent improvement for them in several settings (\ie, few-shot learning, cross-domain generalization, cross-dataset generalization, \etc) over 11 datasets. Further, experiments show that GRAM enables the orthogonal methods of textual and visual prompt tuning to work in a mutually-enhanced way, offering better generalizability beyond the uni-modal prompt tuning methods.

\end{abstract}


\section{Introduction}

Pre-trained on vast image-text pairs that cover almost an infinite range of concepts in the real-world, recent vision-language pre-training models~\cite{radford2021learning, jia2021scaling, li2022fine} have exhibited impressive generalizability on a wide variety of downstream tasks~\cite{antol2015vqa, lin2014microsoft, li2022end, li2023winner, zhang2022magic}. By simply infilling a hand-crafted prompt template (\eg, ``\verb|a photo of a [CLASS]|'') with real class names as input to the text encoder, the pre-training models can achieve zero-shot image classiﬁcation. While effective, a slight word change in prompt templates could lead to a huge diﬀerence in performance~\cite{zhou2022learning}. Thus, identifying suitable prompts for different tasks requires time-consuming attempts by experts on an extra large validation set. Instead of manually designing hard prompts (discrete language words), some recent prompt tuning methods~\cite{zhou2022learning, zhou2022conditional, zhu2022prompt, jia2022visual, lester2021power} are proposed to learn a set of soft prompts (continuous embeddings) using a few labeled data.

\begin{figure}[!t]
	\centering
	\includegraphics[width=\linewidth]{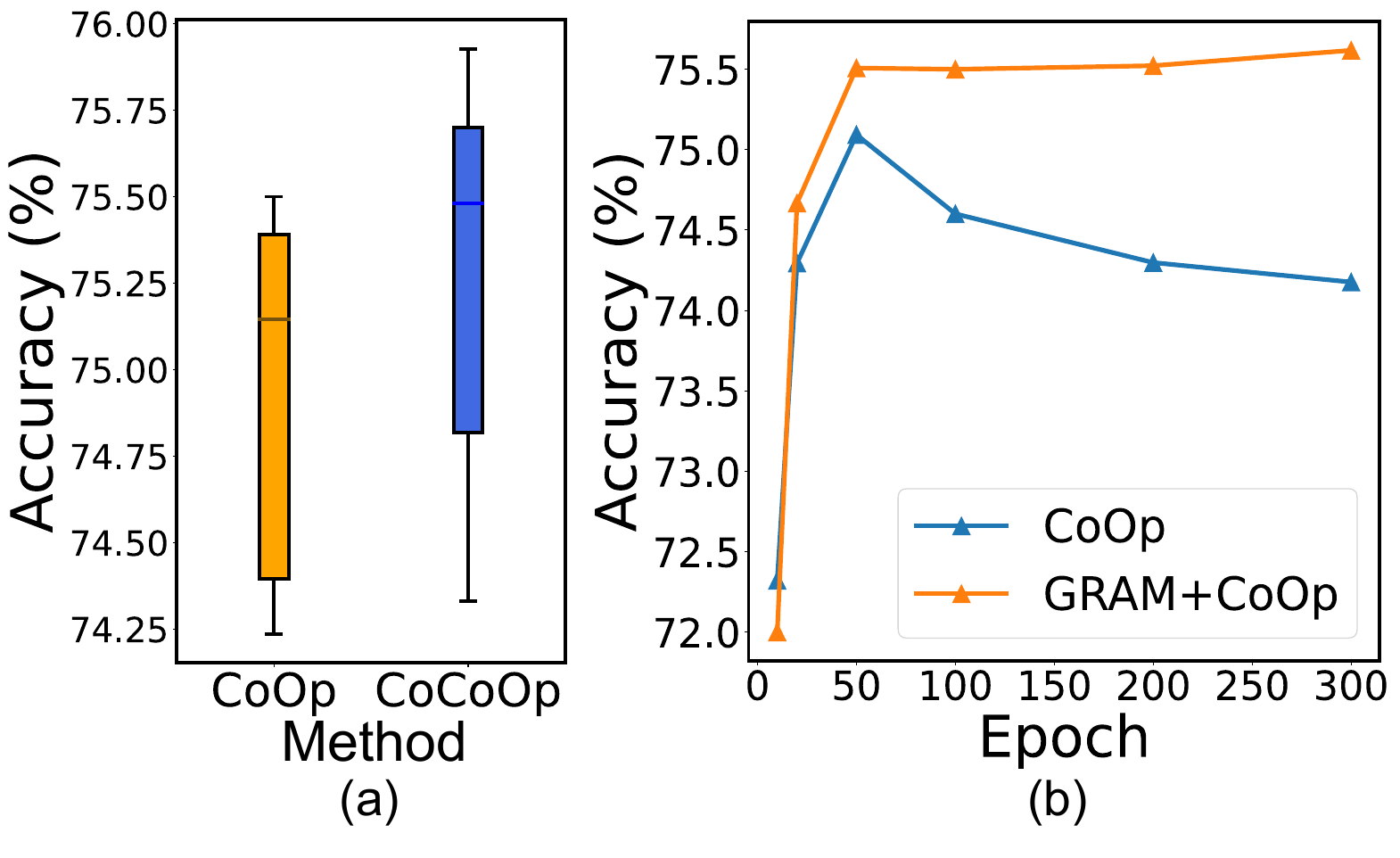}
	\vspace{-0.8cm}
	\caption{(a) Prompt tuning accuracy varies significantly with different initialization. (b) As the training continues, CoOp's performance drops severely while our GRAM prevents CoOp from overfitting to spurious correlations.}
	\label{motivation}
	\vspace{-0.5cm}
\end{figure}

Despite clear improvements on the downstream tasks, prompt tuning for few-shot generalization still has two limitations: (1) \textbf{Initialization-sensitive issue:} performance is particularly sensitive to the initialization of soft prompts. Figure~\ref{motivation}(a) shows that the average few-shot performance varies significantly due to different initialization. Every time we encounter a new task, we need to carefully tune different initialization, which restricts the pre-training models from fast adapting to new tasks. (2) \textbf{Generalizability degradation:} since all the prompt tokens are fine-tuned on limited training samples, it can easily overfit to some spurious correlations or in-distribution patterns, damaging the generalizability of the pre-training models. As shown in Figure~\ref{motivation}(b), CoOp achieves the best results at the early stage. However, as the training continues, its generalizability decreases significantly.

In this paper, we propose a novel Gradient-RegulAted Meta-prompt learning~(GRAM) framework to jointly meta-learn \textbf{an efficient soft prompt initialization} that learns to better adapt to new prompting tasks and \textbf{a lightweight gradient regulating function} that learns to transform the raw fine-tuning gradient into a consistent direction across domains to prevent prompt tuning from damaging the generalizability of the pre-training models. 

Meta-learning~\cite{finn2017model}, also known as learning to learn, optimizes the ability to quickly learn new tasks with only a few samples by transferring the knowledge from learning across a set of meta-training tasks. Typical meta-learning algorithms usually assume access to a distribution of well-annotated meta-training tasks. Differently, we resort to large-scale image-text pairs on the Internet, which is easily available and contains a broader set of visual concepts.

Specifically, we first design a Cross-Modal Hierarchical Clustering algorithm to organize the large-scale image-text data into a hierarchical structure, where the image-text data is first grouped into different semantic topics according to the text descriptions, and each topic of data is further grouped into multiple domains according to the image contents. Then, a diverse set of meta-training classification tasks can be derived by subsampling from the set of semantic topics. For each meta-training task, we simulate domain shift between support set and query set by sampling examples from different domains. The meta-optimization objective is then defined as: after fine-tuning the prompt initialization by one or a few steps using the regulated gradient over a few support set samples, the newly prompted pre-training model should directly perform well on the query set domain. The soft prompt initialization and the gradient regulating function are jointly updated according to the meta gradient directions over the query set samples, thus explicitly learning to better adapt to the new tasks and to avoid overfitting to specific in-domain biases. 

Moreover, we provide analysis to show that the proposed gradient regulating function is learned to regulate the gradient into a consistent direction across domains, thus avoiding overfitting to some spurious correlations of a single domain. Note that, our method is model-agnostic. Comprehensive experiments show that GRAM is generalizable to different prompt tuning methods, significantly boosting all models' performance and generalizability. Further, GRAM enables the harmonious and efficient integration of two orthogonal methods - textual prompt tuning~(\ie, CoOp) and visual prompt tuning~(\ie, VPT). By jointly meta-learning an efficient initialization for both textual and visual prompts, GRAM ensures that both the textual and visual prompts are optimized for better adaptation to new tasks in a complementary way. The resulting UNIversal Gradient-RegulAted Meta-prompt~(UNIGRAM) leverages this seamless integration to unlock the greater potential of both methods and achieves superior few-shot generalization performance. Our contributions are mainly three-fold:

\begin{itemize}
	\vspace{-0.2cm}
	\item We propose an innovative Gradient-RegulAted Meta-prompt learning~(GRAM) framework that explicitly optimizes the adaptation capability to new prompting tasks and the generalization capability to novel domains in a bi-level meta-learning paradigm using only unlabeled image-text pre-training data. 
	\vspace{-0.1cm}
	\item GRAM can be easily incorporated into different prompt tuning methods in a plug-and-play fashion, and the extensive experiments over 11 datasets illustrate the superior generalizability of our GRAM on base-to-new, cross-domain, and cross-dataset generalization.
	\vspace{-0.1cm}
	\item In addition, GRAM enables the orthogonal methods of textual and visual prompt tuning to work in a mutually-enhanced manner, offering stronger generalizability.

\end{itemize}

\section{Related Work}

\noindent
\textbf{Prompt Tuning.} Prompt tuning is first introduced in the NLP area~\cite{radford2019language} to close the gap between pre-training and downstream tasks. Petroni~\textsl{et al.}~\cite{petroni2019language} manually create cloze-style prompts to elicit knowledge from pre-trained language models in a ``ﬁll-in-the-blank'' way. Further, prompt tuning is introduced in vision-language understanding~\cite{zhou2022learning}, which can enhance the generalizability of large vision-language models~\cite{radford2021learning, jia2021scaling, li2023empowering} on a wide range of vision-language understanding tasks~\cite{zhang2019frame, ji2023binary, li2022compositional, li2020topic, li2021adaptive, yu2023visually, ji2023partial, zhang2022boss, li2022hero}.
As manually designing suitable prompts for different tasks is time-consuming and usually sub-optimal, recent works~\cite{zhou2022learning, jia2022visual, lester2021power} propose to optimize a set of continuous learnable prompt embeddings. Concretely, CoOp~\cite{zhou2022learning} optimizes continuous prompt embeddings to improve the few-shot generalizability of CLIP. CoCoOp~\cite{zhou2022conditional} proposes to learn image-conditioned prompts to further improve the generalizability of CoOp. ProDA~\cite{lu2022prompt} learns a distribution of diverse prompts via Gaussian distribution to handle the varying visual representations. To further enhance CLIP’s adaption capability, Tip-Adapter builds a key-value cache model from the few-shot training samples to perform feature retrieval. Instead of designing a specific prompt tuning method, we propose a model-agnostic meta-prompt learning framework to improve the adaptation ability and cross-domain generalizability of the prompt tuning methods, which can be incorporated into existing methods in a plug-and-play fashion.

\noindent
\textbf{Meta-Learning.} Meta-learning aims to enable efficient adaptation ability of models by leveraging the experience from learning across a set of tasks. Meta-learning approaches can be categorized as: \textsl{metric-based}~\cite{snell2017prototypical, sung2018learning, vinyals2016matching}, \textsl{memory-based}~\cite{mishra2017simple, munkhdalai2017meta, oreshkin2018tadam, santoro2016meta}, and \textsl{optimization-based}~\cite{nichol2018first, finn2017model, hochreiter2001learning, ravi2016optimization, frans2017meta, li2020unsupervised}. Our framework is based on the optimization-based method~(\ie, MAML~\cite{finn2017model}). Rather than relying on human-annotated meta-training tasks, our method can automatically generate a diverse set of meta-training tasks by cross-modal hierarchical clustering. Li~\textsl{et al.}~\cite{li2018learning} propose to synthesize domain shift during meta-training to learn a domain-generalizable initialization. Differently, we present a novel gradient regulating function that actively transforms the updated gradient into a domain-generalizable direction.

\section{Method}

In this section, we first introduce the preliminaries in Section~\ref{s3.1}. Next, we present the Cross-Modal Hierarchical Clustering to automatically construct a diverse set of meta-training tasks in Section~\ref{s3.2}. Then, in Section~\ref{s3.3}, we elaborate on how GRAM jointly meta-learns an efficient soft prompt initialization and a lightweight gradient regulating function from these meta-training tasks. 
Finally, we provide theoretical analysis to better understand how our GRAM can improve generalizability in Section~\ref{s3.4}.

\vspace{-0.1cm}
\subsection{Preliminaries}~\label{s3.1}
\vspace{-0.4cm}

\noindent
\textbf{Contrastive Language-Image Pre-Training.} CLIP~\cite{radford2021learning} aims to learn an image encoder $f_I$ and a text encoder $f_T$ by contrastive language-image pre-training paradigm on tremendous image-text pairs, where the matched image-text pairs are optimized to get closer in the joint semantic space. After pre-training, CLIP can generalize to zero-shot visual recognition by reformulating classification as an image-text matching problem. Concretely, the ``\verb|[CLASS]|'' name can be extended to an input sentence to the text encoder $f_T$ by filling a prompt template like ``\verb|a photo of a [CLASS]|''. Let $f_T(\mathbf{T}_i)$  denotes the class-extended text feature for the $i-$th class, and then the probability for the $i-$th class is defined as:

\vspace{-0.3cm}
\begin{equation}
	\vspace{-0.2cm}
	p(y = i | \mathbf{I}) = \frac{\exp (\mathrm{sim}(f_T(\mathbf{T}_i), f_I(\mathbf{I}))/\tau)}{\sum_{j=1}^{J} \exp (\mathrm{sim}(f_T(\mathbf{T}_j), f_I(\mathbf{I}))/\tau)}
\end{equation}

\noindent
where $\mathrm{sim(\cdot, \cdot)}$ denotes cosine similarity, $J$ is the number of classes, and $\tau$ is a learned temperature parameter.

\noindent
\textbf{Textual Prompt Tuning.} To avoid the time-consuming process of identifying customized prompts for different tasks, Context Optimization~(CoOp)~\cite{zhou2022learning} proposes to learn a set of  $M$ continuous prompt vectors $\{\mathbf{t}_1, \mathbf{t}_2, ..., \mathbf{t}_M\}$ to replace the hand-craft prompt template. The prompt sentence for the $i$-th class is constructed by concatenating the prompt vectors with the word embedding of the class name $c_i$: 

\vspace{-0.2cm}
\begin{equation}
	\vspace{-0.1cm}
	\hat{\mathbf{T}}_i = [\mathbf{t}_1, \mathbf{t}_2, ..., \mathbf{t}_M, \mathbf{c}_i]
\end{equation}

\noindent
In downstream tasks, since the pre-training model is frozen, the learnable prompt vectors can be efficiently optimized by minimizing the cross-entropy loss using only a few samples.

\noindent
\textbf{Visual Prompt Tuning.} Recently, similar prompt tuning ideas~\cite{bahng2022visual, jia2022visual} have been proposed for the vision Transformer encoder, where a set of  learnable prompt vectors are concatenated with the input image patch tokens, with the goal of extracting more transferable visual features.

\begin{figure*}[!t]
	\centering
	\includegraphics[width=\textwidth]{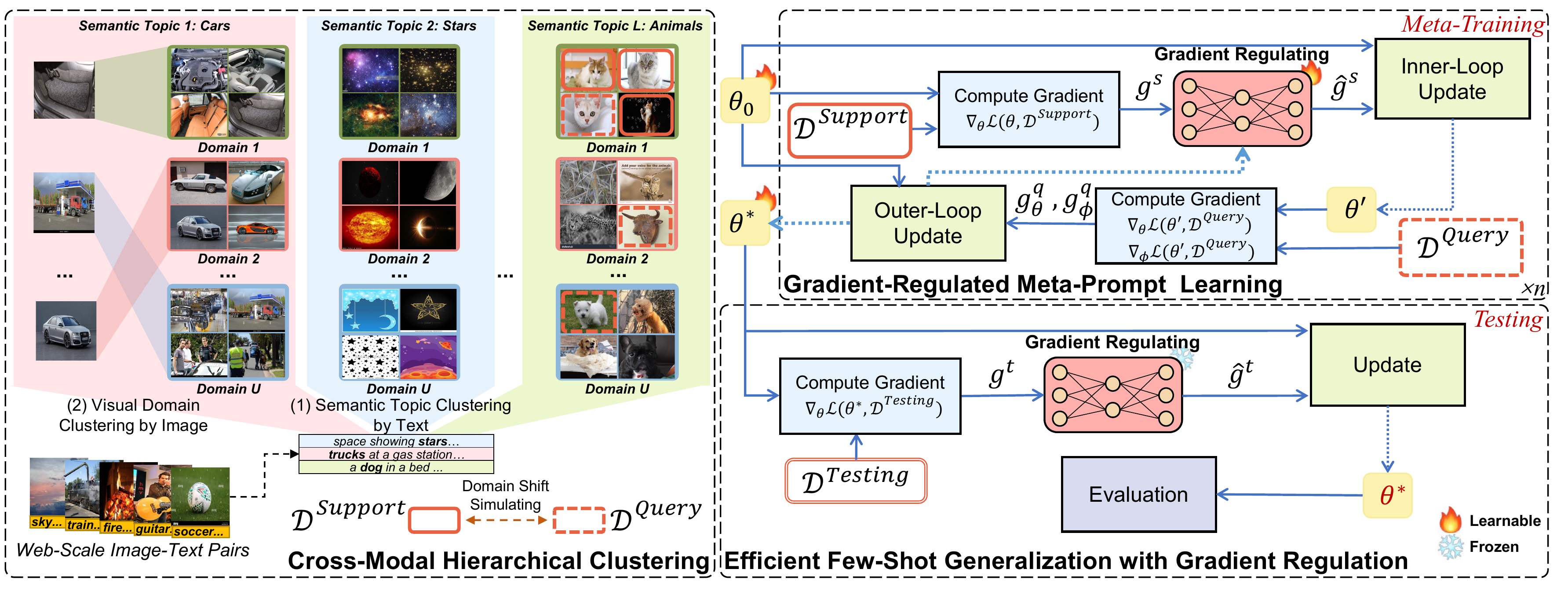}
	\vspace{-0.7cm}
	\caption{Overview of the proposed GRAM framework.}
	\label{framework}
	\vspace{-0.3cm}
\end{figure*}

\subsection{Cross-Modal Hierarchical Clustering}~\label{s3.2}
\vspace{-0.3cm}

As large-scale image-text pairs are easily available on the Internet and cover more comprehensive semantic concepts than any existing human-annotated dataset, we present a Cross-Modal Hierarchical Clustering~(CHC) algorithm to construct a diverse and structured set of meta-training tasks from the image-text pairs. As shown in Figure~\ref{framework}, CHC organizes the image-text pairs into a hierarchical structure through two steps: \textbf{semantic topic clustering} and \textbf{visual domain clustering}. The semantic topic clustering first groups image-text pairs into different clusters according to their text descriptions, where each cluster corresponds to a semantic topic. Next, the visual domain clustering further partitions the data in each of the semantic topics into consistent and distinct subsets based on their image features.

\noindent
\textbf{Semantic Topic Clustering.} To group image-text data based on their underlying semantics, we employ BERTopic~\cite{grootendorst2022bertopic} to cluster the text descriptions. Specifically, for each image-text pair, we first use Sentence-BERT~\cite{reimers2019sentence} to encode the text sentence into a dense sentence embedding. To avoid the semantic space collapse problem where the spatial locality becomes ill-deﬁned and distance measures differ little in high dimensional space, we adopt UMAP~\cite{mcinnes2018umap} to reduce the dimensionality of sentence embeddings while preserving the local and global features of high-dimensional data. Then, we cluster the reduced embeddings by the standard clustering algorithm HDBSCAN~\cite{mcinnes2017hdbscan}.

After obtaining $L$ clusters of image-text data $\mathcal{P} = \{\mathcal{C}^l\}_{l=1}^L$, we extract the semantic topic word for each cluster $\mathcal{C}^l$ according to a cluster-wise TF-IDF~\cite{joachims1996probabilistic}, which measures the importance of a word to a cluster. Specifically, we treat all text sentences in a cluster as a single document by concatenating the sentences. Then, the cluster-wise TF-IDF score for word $w$ in the cluster $\mathcal{C}^l$ is defined as:

\vspace{-0.2cm}
\begin{equation}
	\vspace{-0.2cm}
	\mathrm{TF-IDF}_{w, l} = \frac{N_{w, l}}{N_l} \cdot \mathrm{log}(\frac{L}{L_w + 1})
\end{equation}

\noindent
where $N_{w, l}$ is the number of times that word $w$ occurs in cluster $\mathcal{C}^l$, $N_l$ is the total number of  words in cluster $\mathcal{C}^l$, $L$ is the number of clusters, and $L_w$ is the number of clusters that contain word $w$. The first term models the frequency of word $w$ in cluster $\mathcal{C}^l$, and the second term measures how much information word $w$ provides. 

Thus, the cluster-wise TF-IDF allows us to extract the most representative word as the semantic topic for each cluster by choosing the word with the highest TF-IDF score.

\noindent
\textbf{Visual Domain Clustering.} After grouping image-text pairs into multiple semantic topics, we perform visual domain clustering to partition each semantic topic of data into several consistent and distinct domains based on the image features. Specifically, we extract the image features using the pre-trained vision encoder.
Next, we run k-means clustering to further group each topic of image-text pairs $\mathcal{C}^l \in \mathcal{P}$ into several domains: $\mathcal{C}^l = \{\mathcal{H}^l_u\}_{u=1}^U$, where $\mathcal{H}^l_u = \{(\mathbf{I}_i, Y^l)\}_{i=1}^{N^l_u}$. Here we omit the paired text of image $\mathbf{I}_i$, and $Y^l$ is the selected semantic topic word for cluster $\mathcal{C}^l$. Thus, we obtain a hierarchical structure of image-text pairs, facilitating to construct diverse meta-training tasks and simulate domain shift during meta-training process. 

\vspace{-0.1cm}
\subsection{Gradient-Regulated Meta-Prompt Learning}~\label{s3.3}
\vspace{-0.4cm}

As shown in Figure~\ref{framework}, GRAM is a bi-level meta-learning paradigm that jointly meta-learns an efficient soft prompt initialization $\theta$ for better adaptation and a lightweight gradient regulating function $R^{\phi}$ to prevent prompt tuning from damaging the generalizability of the pre-training models. As GRAM is a model-agnostic method, $\theta$ can represent the parameters of any type of prompt tuning method. 

\noindent
\textbf{Automatic Meta-Training Task Generation.} To construct a $K_t$-way image classification task $\tau_t$, we first sample $K_t$ clusters from $\mathcal{P} = \{\mathcal{C}^l\}_{l=1}^L$. Each sampled cluster corresponds to a category of images with the class label $Y^l$. Then, we sample a few image instances from the selected clusters to construct the support set $\mathcal{D}^{\mathrm{support}}_t$ and query set $\mathcal{D}^{\mathrm{query}}_t$ for the meta-training task $\tau_t$. Note that, for the support set, we restrict the images to be only sampled from a single domain of each selected cluster. As for the query set, we uniformly sample more images from all domains of each selected cluster. In this way, we \textbf{simulate the train/test domain shift} during meta-training. The support set samples are \textbf{domain-specific} and the query set samples are \textbf{more representative}. Following the above procedure, we construct a diverse set of meta-training tasks $\mathcal{T} = \{\tau_t\}_{t=1}^T$. 

\noindent
\textbf{Meta-Training Overview.} Our bi-level meta-learning paradigm mainly consists of two optimization steps. In the \textit{inner-loop}, the initialization $\theta$ is adapted to each meta-training task $\tau_t$ according to the gradient regulated by $R^{\phi}$ over a few support set samples $\mathcal{D}^{\mathrm{support}}_t$, and then, in the \textit{outer-loop}, a meta-learning objective evaluates the adaptation and generalization capabilities of the adapted model on a distinguished query set $\mathcal{D}^{\mathrm{query}}_t$. The initialization $\theta$ and gradient regulating function $R^{\phi}$ are jointly optimized according to the performance of the adapted model on the query set across a wide range of meta-training tasks.

\noindent
\textbf{Inner-Loop.} Formally, we consider adapting $\theta$ to a new task $\tau_t$. In the inner-loop, the prompt parameters are first updated via gradient descent over support set $\mathcal{D}^{\mathrm{support}}_t$:

\vspace{-0.2cm}
\begin{equation}
	\theta_{t}' \longleftarrow \theta - \alpha\nabla_{\theta}\mathcal{L}(\theta, \mathcal{D}^{\mathrm{support}}_t) \label{e4}
\end{equation}

\noindent
where $\mathcal{L}$ and $\alpha$ denote the loss function and the inner-loop learning rate, respectively. 

While straightforward, the above update step on limited samples might overfit to some domain-specific patterns, undermining the generalizability of the pre-training models on other domains. Thus, we propose to meta-learn an effective and efficient \textbf{gradient regulating function}, which can transform the raw gradient into a more consistent direction across domains while ignoring spuriously correlations. 

Concretely, gradient regulating function $R^{\phi}$ parameterized by $\phi$ performs affine transformation to modulate the raw gradient for generalizable fine-tuning. Given $\mathbf{g}_t = \nabla_{\theta}\mathcal{L}(\theta, \mathcal{D}^{\mathrm{support}}_t) \in \mathbb{R}^{d \times M}$ as input, $R^{\phi}$ first generates two modulation vectors $\gamma_t \in \mathbb{R}^{d \times M}$ and $\beta_t \in \mathbb{R}^{d \times M}$ as follows:

\vspace{-0.4cm}
\begin{equation}
	\vspace{-0.1cm}
	\gamma_t = \mathrm{tanh}(\mathbf{W}^{\gamma} \mathbf{g}_t + \mathbf{b}^{\gamma}),\quad  \beta_t = \mathrm{tanh}(\mathbf{W}^{\beta} \mathbf{g}_t + \mathbf{b}^{\beta})
\end{equation}

\noindent
where $\mathbf{W}^{\gamma}, \mathbf{b}^{\gamma}, \mathbf{W}^{\beta}$ and $\mathbf{b}^{\beta}$ are learnable parameters. Then, the raw gradient $\mathbf{g}_t$ is regulated as:
\vspace{-0.2cm}
\begin{equation}
	\vspace{-0.1cm}
	\hat{\mathbf{g}}_t = \gamma \odot \mathbf{g}_t  + \beta_t 
\end{equation}

Consequently, Equation~\ref{e4} can be transformed as:

\vspace{-0.2cm}
\begin{equation}
	\theta_{t}' \longleftarrow \theta - \alpha R^{\phi}(\nabla_{\theta}\mathcal{L}(\theta, \mathcal{D}^{\mathrm{support}}_t))\label{e7}
\end{equation}

\noindent
\textbf{Outer-Loop.} After adapting the soft prompt parameters to the task $\tau_t$ according to the support set $\mathcal{D}^{\mathrm{support}}_t$, the initialization parameters $\theta$ and the parameters of the gradient regulating function $\phi$ are optimized for the performance of the adapted parameters $\theta'$ on the query set $\mathcal{D}^{\mathrm{query}}_t$:

\vspace{-0.4cm}
\begin{align}
	\vspace{-0.2cm}
	\theta &\longleftarrow \theta - \lambda_1 \nabla_{\theta}\mathcal{L}(\theta_{t}', \mathcal{D}^{\mathrm{query}}_t)\\
	\phi &\longleftarrow \phi - \lambda_2 \nabla_{\phi}\mathcal{L}(\theta_{t}', \mathcal{D}^{\mathrm{query}}_t)
\end{align}

\noindent
where $\lambda$ denotes the outer-loop learning rate. Overall, the meta-optimization objective can be formulated as:

\vspace{-0.3cm}
\begin{equation}
	\vspace{-0.1cm}
	\mathrm{min}_{\theta, \phi} \sum_{\tau_t \in \mathcal{T}} \mathcal{L}(\theta - \alpha R^{\phi}(\nabla_{\theta}\mathcal{L}(\theta, \mathcal{D}^{\mathrm{support}}_t)), \mathcal{D}^{\mathrm{query}}_t) \label{e10}
\end{equation}

\noindent
The meta-optimization is performed across a wide range of meta-training tasks. Thus, the initialization $\theta$ is explicitly optimized to better adapt to new tasks and $R^{\phi}$ is optimized to transform the raw gradient so that the model can generalize to the unseen domains of the query sets. The overall methodological ﬂow is summarized in Algorithm~\ref{A1}.

\noindent
\textbf{Testing.} At test-time, the optimized $\theta^*$ is deployed as the soft prompt initialization and adapted to testing tasks using the regulated gradient over few-shot samples as Equation~\ref{e7}.

\noindent
\textbf{Universal Gradient-Regulated Meta-Prompt.} As a model-agnostic approach, our later experimental results show that GRAM can significantly boost both textual and visual prompt tuning, respectively. Based on this observation, we further present UNIversal Gradient-RegulAted Meta-prompt~(UNIGRAM) to explore whether our GRAM enables the visual and textual prompts to cooperate in a complementary way. Without any architecture modification, we meta-learn an effective initialization $\theta = [\theta_T, \theta_V]$ and a corresponding gradient regulating function, where $\theta_T$ and $\theta_V$ represent the parameters of textual prompt vectors and visual prompt vectors, respectively.

\begin{table*}[t]
	\tabstyle{6pt}
	\caption{Accuracy (\%) of  base-to-new generalization evaluation over 11 datasets. Prompts are learned from the base classes (16 shots). H: Harmonic mean, which is used to highlight the generalization trade-off. The best results are highlighted in \color{WildStrawberry}{red}.}
	\vspace{-0.2cm}
	\label{t1}
	\begin{subtable}[t]{.32\textwidth}
		\centering
		\caption{\textbf{Average over 11 datasets}.}
		\vspace{-0.2cm}
		\begin{tabular}{l cc|c}
			\toprule
			& Base & New & H \\
			\hline
			CLIP & 69.34 & 74.22 & 71.70 \\
			CoCoOp &\color{WildStrawberry}{80.47} & 71.69 &75.83 \\
			\hline
			CoOp    &77.58  &73.11  &75.28   \\
			\textbf{+ GRAM}   &\textbf{78.74}  &\textbf{74.93}  &\textbf{76.79}   \\
			\hline
			VPT   &72.53  &72.34  &72.43   \\
			\textbf{+ GRAM}   &\textbf{74.04}  &\textbf{74.21}  &\textbf{74.12}   \\
			\hline
			\textbf{UNIGRAM}  &80.34   & \color{WildStrawberry}{75.92} &\color{WildStrawberry}{78.07} \\
			\bottomrule
		\end{tabular}
	\end{subtable}
	\vspace{0.5em}
	\begin{subtable}[t]{.32\textwidth}
		\centering
		\caption{ImageNet.}
		\vspace{-0.2cm}
		\begin{tabular}{l cc|c}
			\toprule
			& Base & New & H \\
			\hline
			CLIP & 72.43 & 68.14 & 70.22 \\
			CoCoOp & 75.98 &70.43 &73.10 \\
			\hline
			CoOp    &76.21 &69.98   &72.96   \\
			\textbf{+ GRAM}   &\textbf{76.42}  &\textbf{70.17} &\textbf{73.16}   \\
			\hline
			VPT   &74.45  &69.22  &71.74   \\
			\textbf{+ GRAM}   &\textbf{74.76}  &\textbf{69.54}  &\textbf{72.06}   \\
			\hline
			\textbf{UNIGRAM}  &\color{WildStrawberry}{76.60}   & \color{WildStrawberry}{70.69} &\color{WildStrawberry}{73.53} \\
			\bottomrule
		\end{tabular}
	\end{subtable}
	~
	\begin{subtable}[t]{.32\textwidth}
		\centering
		\caption{Caltech101.}
		\vspace{-0.2cm}
		\begin{tabular}{l cc|c}
			\toprule
			& Base & New & H \\
			\hline
			CLIP & 96.84 &94.00 & 95.40 \\
			CoCoOp & 97.96 & 93.81 & 95.84 \\
			\hline 
			CoOp     &97.49   &94.67   &96.06   \\
			\textbf{+ GRAM}   &\textbf{98.07} &\textbf{95.00}  &\textbf{96.51}   \\
			\hline
			VPT   &96.92  &93.44  &95.15   \\
			\textbf{+ GRAM}  &\textbf{97.33}  &\textbf{94.11}  &\textbf{95.69}    \\
			\hline
			\textbf{UNIGRAM}  &\color{WildStrawberry}{98.07}   & \color{WildStrawberry}{95.11} &\color{WildStrawberry}{96.57}   \\
			\bottomrule
		\end{tabular}
	\end{subtable}
	~
	\begin{subtable}[t]{.32\textwidth}
		\centering
		\caption{OxfordPets.}
		\vspace{-0.2cm}
		\begin{tabular}{l cc|c}
			\toprule
			& Base & New & H \\
			\hline
			CLIP & 91.17 & 97.26 & 94.12 \\
			CoCoOp &\color{WildStrawberry}{95.20} & 97.69 &\color{WildStrawberry}{96.43} \\
			\hline
			CoOp    &94.48  & 95.88 &95.17   \\
			\textbf{+ GRAM}   &\textbf{94.71}  &\textbf{97.52}  &\textbf{96.09}   \\
			\hline
			VPT   &92.63  &94.96  &93.78   \\
			\textbf{+ GRAM}   &\textbf{93.50}  &\textbf{97.06}  &\textbf{95.25}    \\
			\hline
			\textbf{UNIGRAM}  &94.94   & \color{WildStrawberry}{97.94} &96.42   \\
			\bottomrule
		\end{tabular}
	\end{subtable}
	\vspace{0.5em}
	\begin{subtable}[t]{.32\textwidth}
		\centering
		\caption{StanfordCars.}
		\vspace{-0.2cm}
		\begin{tabular}{l cc|c}
			\toprule
			& Base & New & H \\
			\hline
			CLIP & 63.37 & 74.89 & 68.65 \\
			CoCoOp & 70.49 & 73.59 & 72.01 \\
			\hline
			CoOp    &71.26  &73.92  &72.57   \\
			\textbf{+ GRAM}   &\textbf{72.08}  &\textbf{74.80}  &\textbf{73.41}  \\
			\hline
			VPT   &65.06  &74.68  &69.54   \\
			\textbf{+ GRAM}   &\textbf{65.65}  &\textbf{75.10}  &\textbf{70.06}   \\
			\hline
			\textbf{UNIGRAM} &\color{WildStrawberry}{73.50}   & \color{WildStrawberry}{75.38} &\color{WildStrawberry}{74.43}   \\
			\bottomrule
		\end{tabular}
	\end{subtable}
	~
	\begin{subtable}[t]{.32\textwidth}
		\centering
		\caption{Flowers102.}
		\vspace{-0.2cm}
		\begin{tabular}{l cc|c}
			\toprule
			& Base & New & H \\
			\hline
			CLIP & 72.08 & 77.80 & 74.83 \\
			CoCoOp & 94.87 & 71.75 & 81.71 \\
			\hline
			CoOp    &87.87  &74.11  &80.41   \\
			\textbf{+ GRAM}   &\textbf{92.60}  &\textbf{75.64}  &\textbf{83.27}   \\
			\hline
			VPT   &76.23  &71.55  &73.82   \\
			\textbf{+ GRAM}   &\textbf{77.10}  &\textbf{74.64}  &\textbf{75.85}    \\
			\hline
			\textbf{UNIGRAM}  &\color{WildStrawberry}{95.20}   & \color{WildStrawberry}{76.21} &\color{WildStrawberry}{84.65}  \\
			\bottomrule
		\end{tabular}
	\end{subtable}
	~
	\begin{subtable}[t]{.32\textwidth}
		\centering
		\caption{Food101.}
		\vspace{-0.2cm}
		\begin{tabular}{l cc|c}
			\toprule
			& Base & New & H \\
			\hline
			CLIP & 90.10 & 91.22 & 90.66 \\
			CoCoOp & 90.70 & 91.29 & 90.99 \\
			\hline
			CoOp    &90.63  &91.17  &90.90   \\
			\textbf{+ GRAM}   &\textbf{90.68}  &\textbf{91.91}  &\textbf{91.29}   \\
			\hline
			VPT   &89.27  &90.50  &89.88   \\
			\textbf{+ GRAM}   &\textbf{89.86}  &\textbf{91.32}  &\textbf{90.58}   \\
			\hline
			\textbf{UNIGRAM}  &\color{WildStrawberry}{90.84}   & \color{WildStrawberry}{92.12} &\color{WildStrawberry}{91.48}  \\
			\bottomrule
		\end{tabular}
	\end{subtable}
	\vspace{0.5em}
	\begin{subtable}[t]{.32\textwidth}
		\centering
		\caption{FGVCAircraft.}
		\vspace{-0.2cm}
		\begin{tabular}{l cc|c}
			\toprule
			& Base & New & H \\
			\hline
			CLIP & 27.19 & 36.29 & 31.09 \\
			CoCoOp &\color{WildStrawberry}{33.41} & 23.71 & 27.74 \\
			\hline
			CoOp    &30.66  &35.73  &33.00   \\
			\textbf{+ GRAM}   &\textbf{31.19}  &\textbf{36.50}  &\textbf{33.64}   \\
			\hline
			VPT   &28.23  &32.21  &30.09   \\
			\textbf{+ GRAM}   &\textbf{28.81}  &\textbf{34.50}  &\textbf{31.40}    \\
			\hline
			\textbf{UNIGRAM}  &32.25   & \color{WildStrawberry}{38.00} &\color{WildStrawberry}{34.89}  \\
			\bottomrule
		\end{tabular}
	\end{subtable}
	~
	\begin{subtable}[t]{.32\textwidth}
		\centering
		\caption{SUN397.}
		\vspace{-0.2cm}
		\begin{tabular}{l cc|c}
			\toprule
			& Base & New & H \\
			\hline
			CLIP & 69.36 & 75.35 & 72.23 \\
			CoCoOp & 79.74 & 76.86 & 78.27\\
			\hline
			CoOp    &79.78  &76.04  &77.87   \\
			\textbf{+ GRAM}   &\textbf{80.09}  &\textbf{76.97}  &\textbf{78.50}   \\
			\hline
			VPT   &75.14  &76.89  &76.00   \\
			\textbf{+ GRAM}   &\textbf{75.74}  &\textbf{77.64}  &\textbf{76.68} \\
			\hline
			\textbf{UNIGRAM}  &\color{WildStrawberry}{80.43}   & \color{WildStrawberry}{77.91} &\color{WildStrawberry}{79.15}   \\
			\bottomrule
		\end{tabular}
	\end{subtable}
	~
	\begin{subtable}[t]{.32\textwidth}
		\centering
		\caption{DTD.}
		\vspace{-0.2cm}
		\begin{tabular}{l cc|c}
			\toprule
			& Base & New & H \\
			\hline
			CLIP & 53.24 & 59.90 & 56.37 \\
			CoCoOp &\color{WildStrawberry}{77.01} & 56.00 & 64.85 \\
			\hline
			CoOp    &69.46  &55.67  &61.81   \\
			\textbf{+ GRAM}   &\textbf{72.87}  &\textbf{59.49}  &\textbf{65.50}   \\
			\hline
			VPT   &56.71  &57.25  &56.98   \\
			\textbf{+ GRAM}  &\textbf{58.25}  &\textbf{58.00}  &\textbf{58.12}  \\
			\hline
			\textbf{UNIGRAM}  &73.62   & \color{WildStrawberry}{62.38} &\color{WildStrawberry}{67.56}   \\
			\bottomrule
		\end{tabular}
	\end{subtable}
	~
	\begin{subtable}[t]{.32\textwidth}
		\centering
		\caption{EuroSAT.}
		\vspace{-0.2cm}
		\begin{tabular}{l cc|c}
			\toprule
			& Base & New & H \\
			\hline
			CLIP & 56.48 & 64.05 & 60.03 \\
			CoCoOp & \color{WildStrawberry}{87.49} & 60.04 & 71.21 \\
			\hline
			CoOp    &74.79  &61.50  &67.50   \\
			+ GRAM  &\textbf{76.00}  &\textbf{69.92}  &\textbf{72.83}   \\
			\hline
			VPT   &67.57  &59.69  &63.39   \\
			+ GRAM  &\textbf{77.26}  &\textbf{68.26}  &\textbf{72.48}  \\
			\hline
			\textbf{UNIGRAM} &86.26   & \color{WildStrawberry}{71.38} &\color{WildStrawberry}{78.12}  \\
			\bottomrule
		\end{tabular}
	\end{subtable}
	~
	\begin{subtable}[t]{.32\textwidth}
		\centering
		\caption{UCF101.}
		\vspace{-0.2cm}
		\begin{tabular}{l cc|c}
			\toprule
			& Base & New & H \\
			\hline
			CLIP & 70.53 & 77.50 & 73.85 \\
			CoCoOp & \color{WildStrawberry}{82.33} & 73.45 & 77.64 \\
			\hline
			CoOp    &80.72  &75.55  &78.05   \\
			\textbf{+ GRAM}   &\textbf{81.47}  &\textbf{76.33}  &\textbf{78.82}   \\
			\hline
			VPT   &75.65  &75.31  &75.48   \\
			\textbf{+ GRAM}   &\textbf{76.21}  &\textbf{76.17}  &\textbf{76.19}  \\
			\hline
			\textbf{UNIGRAM}  &82.00   & \color{WildStrawberry}{78.06} &\color{WildStrawberry}{79.98}   \\
			\bottomrule
		\end{tabular}
	\end{subtable}
	\vspace{-0.3cm}
\end{table*}
\subsection{How GRAM Improves Generalizability}~\label{s3.4}
\vspace{-0.3cm}

In this subsection, we analyze formally how our GRAM can improve generalizability. Let us consider the ﬁrst order Taylor expansion of  the meta-optimization objective at a point $\mathbf{x}_0$ (we omit the subscript $t$ for clarity):
\vspace{-0.2cm}
\begin{equation}
	\mathcal{L}(\mathbf{x}, \mathcal{D}^{\mathrm{query}}) = \mathcal{L}(\mathbf{x}_{0}, \mathcal{D}^{\mathrm{query}}) + \nabla_{\mathbf{x_0}} \mathcal{L}(\mathbf{x}_{0}, \mathcal{D}^{\mathrm{query}}) \cdot (\mathbf{x} - \mathbf{x}_{0})
\end{equation}

Assume we have $\mathbf{x} = \theta - \alpha R^{\phi}(\nabla_{\theta}\mathcal{L}(\theta, \mathcal{D}^{\mathrm{support}}_t))$ and $\mathbf{x}_{0} = \theta$. Then, Equation~\ref{e10} can be reformulated as:
\vspace{-0.1cm}
	\begin{align}
		\begin{aligned}
		\mathrm{min}&_{\theta, \phi}\quad  \mathcal{L}(\mathbf{x}, \mathcal{D}^{\mathrm{query}}) =  \mathrm{min}_{\theta, \phi}\quad \mathcal{L}(\theta, \mathcal{D}^{\mathrm{query}})\\
			&- \alpha R^{\phi}(\nabla_{\theta}\mathcal{L}(\theta, \mathcal{D}^{\mathrm{support}})) \cdot \nabla_{\theta} \mathcal{L}(\theta, \mathcal{D}^{\mathrm{query}})
		\end{aligned}
	\end{align}

\noindent
where the first term represents the loss on the query set and the second term represents the inner product between the regulated gradient over the support set and the gradient over the query set. We are therefore jointly learning to minimize the loss on the query set and maximize the similarity between the gradients. A high similarity means a ``similar gradient direction'' between the support set domain and the query set domain, which indicates that tuning on the support domain will improve the performance on the query domain.

\begin{figure}[t]
	\vspace{-0.3cm}
	\begin{algorithm}[H]
		\caption{Gradient-Regulated Meta-Prompt Learning}
		\begin{algorithmic}[1]
			\State Randomly initialize $\theta, \phi$
			
			\While{not converged}
			\State Sample a batch of tasks $\{\tau_t\}_{t=1}^B$ from $\mathcal{T}$
			
			\ForAll{${\tau}_{t} = \{\mathcal{D}^{\mathrm{support}}_t, \mathcal{D}^{\mathrm{query}}_t\}$}
			\State Evaluate  $\nabla_{\theta}\mathcal{L}(\theta, \mathcal{D}^{\mathrm{support}}_t)$ on $\mathcal{D}^{\mathrm{support}}_t$
			\State Regulate $\nabla_{\theta}\mathcal{L}(\theta, \mathcal{D}^{\mathrm{support}}_t)$  via $R^{\phi}$
			\State $\theta_{t}' \longleftarrow \theta - \alpha R^{\phi}(\nabla_{\theta}\mathcal{L}(\theta, \mathcal{D}^{\mathrm{support}}_t))$
			\EndFor
			
			\State $\theta \longleftarrow \theta - \lambda_{1} \nabla_{\theta} \sum_{\tau_t} \mathcal{L}(\theta_t', \mathcal{D}^{\mathrm{query}}_t)$
			\State $\phi \longleftarrow \phi - \lambda_{2} \nabla_{\phi} \sum_{\tau_t} \mathcal{L}(\theta_t', \mathcal{D}^{\mathrm{query}}_t)$ 
			\EndWhile
		\end{algorithmic}
		\label{A1}
	\end{algorithm}
	\vspace{-0.6cm}
\end{figure}

Recall that we simulate domain shift between the support set and the query set, where the support set samples are domain-specific while the query set samples are more representative across domains. Thus, the gradient over the query set samples represents a more general direction, which is consistent across domains. To improve the gradient alignment, the parameters of the gradient regulating function $\phi$ are meta-optimized to regulate the raw gradient over the support set into a more generalizable direction, thus avoiding overfitting to some domain-specific correlations.

\section{Experiments}
In this section, we evaluate our approach on three settings: (1) generalization from base to new classes within a dataset~(Section~\ref{base-to-new}); (2) cross-domain generalization~(Section~\ref{cross-domain}); (3) cross-dataset generalization~(Section~\ref{cross-dataset}).  

\subsection{Experimental Setup}
\vspace{-0.2cm}

\noindent
\textbf{Datasets.} For base-to-new generalization and cross-dataset generalization, we use 11 image recognition datasets, which cover a diverse set of recognition tasks: ImageNet~\cite{deng2009imagenet} and Caltech101~\cite{fei2004learning} for generic object recognition; OxfordPets~\cite{parkhi2012cats}, StanfordCars~\cite{krause20133d}, Flowers102~\cite{nilsback2008automated}, Food101~\cite{bossard2014food} and FGVCAircraft~\cite{maji2013fine} for fine-grained classiﬁcation; UCF101~\cite{soomro2012ucf101} for action recognition; SUN397~\cite{xiao2010sun} for scene recognition; DTD~\cite{cimpoi2014describing} for texture classiﬁcation; and EuroSAT~\cite{helber2019eurosat} for satellite imagery classification. For cross-domain generalization, we train our model on ImageNet and evaluate the domain generalizability on four variants of ImageNet: ImageNetV2~\cite{recht2019imagenet}, ImageNetSketch~\cite{wang2019learning}, ImageNet-A~\cite{hendrycks2021natural}, and ImageNet-R~\cite{hendrycks2021many}.

\noindent
\textbf{Baselines.} We use the following baselines: (1) Hand-crafted prompt method: Zero-Shot CLIP~\cite{radford2021learning}; (2) Textual prompt tuning methods: CoOp~\cite{zhou2022learning}, CoCoOp~\cite{zhou2022conditional}; (3) Visual prompt tuning method: VPT~\cite{jia2022visual}.

\begin{table*}[!t]
	\tabstyle{7pt}
	\caption{Accuracy (\%) of  cross-domain generalization evaluation. Prompts are learned from the source dataset (16 shots).}
	\vspace{-0.1cm}
	\label{t2}
	\begin{tabular}{l cccccc}
		\toprule
		& Source & \multicolumn{5}{c}{Target} \\ \cmidrule(lr){2-2} \cmidrule(lr){3-7}
		& ImageNet & ImageNetV2 & ImageNet-Sketch & ImageNet-A & ImageNet-R &Average\\
		\hline
		CLIP  & 66.73 & 60.83 & 46.15 & 47.77 & 73.96  &57.18\\
		CoCoOp  & 71.02 & 64.07 & 48.75 &50.63 &76.18 &59.91\\
		\hline
		CoOp   &71.35          &64.28           &48.67        &50.65            &76.50       &60.03     \\
		\textbf{+ GRAM} &\textbf{71.62}        &\textbf{64.66}           &\textbf{49.06}            &\textbf{51.12}   &\textbf{76.76} &\textbf{60.40} \\
		\hline
		VPT  &  68.92   &61.84  &47.64  &46.50   &75.86    &57.96            \\
		\textbf{+ GRAM}  &\textbf{ 69.09}        &\textbf{62.33}           &\textbf{47.92}            &\textbf{47.13}  &\textbf{76.26} &\textbf{58.41} \\
		\hline
		\textbf{UNIGRAM} & \color{WildStrawberry}{71.65} &\color{WildStrawberry}{64.81} &\color{WildStrawberry}{49.54}   & \color{WildStrawberry}{51.51} &\color{WildStrawberry}{77.34}  &\color{WildStrawberry}{60.80} \\
		\bottomrule 
	\end{tabular}
	\vspace{-0.1cm}
\end{table*}

\begin{table*}[!t]
	\tabstyle{5pt}
	\caption{Accuracy (\%) of  cross-dataset generalization evaluation. Prompts are learned from the source dataset (16 shots).}
	\vspace{-0.1cm}
	\label{t3}
	\begin{tabular}{l c ccccccccccc}
		\toprule
		& Source & \multicolumn{11}{c}{Target} \\ \cmidrule(lr){2-2} \cmidrule(lr){3-13}
		& \rotbox{ImageNet} & \rotbox{Caltech101} & \rotbox{OxfordPets} & \rotbox{StanfordCars} & \rotbox{Flowers102} & \rotbox{Food101} & \rotbox{Aircraft} & \rotbox{SUN397} & \rotbox{DTD} & \rotbox{EuroSAT} & \rotbox{UCF101} & \rotbox{Average} \\
		\midrule

		CoCoOp & 71.02 & 94.43 & 90.14 &65.32 &71.88 & 86.06 &22.94 &67.36 &45.73 &45.37 &68.21 &65.74 \\
		
		\hline
		CoOp  &71.35  &93.60   &89.84  &64.74  &70.83  &85.97  &23.03  &66.16  &44.21  &45.95  &68.65 &65.30\\
		\textbf{+ GRAM} &\textbf{71.62}        &\textbf{94.28}           &\textbf{90.17}            &\textbf{65.76}  &\textbf{71.92} &\textbf{86.33}        &\textbf{23.76}           &\textbf{66.75}            &\textbf{45.75}  &\textbf{51.15} &\textbf{69.16} &\textbf{66.50}\\
		\hline
		VPT &68.92 &93.07  &89.44   &64.77   &67.79  &84.91  &23.72  &66.16  &45.02  &37.74  &67.00 &63.96  \\
		\textbf{+ GRAM} &\textbf{69.09}        &\textbf{93.62}           &\textbf{90.03}            &\textbf{65.56}  &\textbf{68.83} &\textbf{85.32}        &\textbf{24.88}   &\textbf{66.77}  &\textbf{45.69} &\textbf{42.01}  &\textbf{67.65} &\textbf{65.04}  \\
		\hline
		\textbf{UNIGRAM} & \color{WildStrawberry}{71.65} &\color{WildStrawberry}{94.67} &\color{WildStrawberry}{90.83}   & \color{WildStrawberry}{66.78} &\color{WildStrawberry}{73.12} & \color{WildStrawberry}{86.69} &\color{WildStrawberry}{25.27} &\color{WildStrawberry}{67.97}   & \color{WildStrawberry}{48.06} &\color{WildStrawberry}{52.63}  &\color{WildStrawberry}{71.03} &\color{WildStrawberry}{67.71}\\
		\bottomrule
	\end{tabular}
\vspace{-0.2cm}
\end{table*}

\noindent
\textbf{Training Details.} For a fair comparison, all methods use CLIP-ViT-B/16 as the pre-training model, and the number of prompt tokens is set to 4, which has been suggested by~\cite{zhou2022conditional} that a shorter context length can lead to better performance. For our UNIGRAM, we use 2 textual and visual prompt tokens, respectively. In all three settings, we evaluate the 16-shot performance, and all methods follow the same training epochs (\ie, 10 epochs), training schedule, and data augmentation settings in CoCoOp. Considering the results of CoOp reported in~\cite{zhou2022conditional} is obtained by training 200 epochs, which extremely undermines the generalizability of CoOp, we re-train CoOp for 10 epochs using the officially released code and find that fewer training epochs significantly improve the generalizability of CoOp. For CHC, we use CC3M~\cite{sharma2018conceptual}, which consists of  3.1 million image-text pairs. Before clustering, we adopt a filtering model to filter out mismatched noisy image-text pairs. 

\begin{table*}[!t]
	\tabstyle{7pt}
	\caption{Accuracy (\%) of  cross-domain generalization evaluation. Prompts are learned from the source dataset (4 shots).}
	\vspace{-0.1cm}
	\label{ap_t1}
	\begin{tabular}{l cccccc}
		\toprule
		& Source & \multicolumn{5}{c}{Target} \\ \cmidrule(lr){2-2} \cmidrule(lr){3-7}
		& ImageNet & ImageNetV2 & ImageNet-Sketch & ImageNet-A & ImageNet-R &Average\\
		\hline
		CLIP  & 66.73 & 60.83 & 46.15 & 47.77 & 73.96  &57.18\\
		CoCoOp  &70.13 &63.05   &46.48  &49.36 &73.80   &58.17\\
		\hline
		CoOp   &69.86          &62.83           &46.90        &48.98            &74.55       &58.32     \\
		\textbf{+ GRAM} &\textbf{70.49}        &\textbf{63.72}           &\textbf{\color{WildStrawberry}{48.42}}            &\textbf{51.13}   &\textbf{76.39} &\textbf{59.92} \\
		\hline
		VPT    &70.11   &62.66  &46.57  &47.99   &74.26   &57.87           \\
		\textbf{+ GRAM}  &\textbf{70.46}        &\textbf{63.93}           &\textbf{48.32}            &\textbf{49.93}  &\textbf{76.37} &\textbf{59.64} \\
		\hline
		\textbf{UNIGRAM} & \color{WildStrawberry}{70.84} &\color{WildStrawberry}{64.01} &48.29   & \color{WildStrawberry}{51.20} &\color{WildStrawberry}{76.76}  &\color{WildStrawberry}{60.07} \\
		\bottomrule 
	\end{tabular}
\end{table*}

\begin{table*}[!t]
	\tabstyle{5pt}
	\caption{Accuracy (\%) of  cross-dataset generalization evaluation. Prompts are learned from the source dataset (4 shots).}
	\vspace{-0.1cm}
	\label{ap_t2}
	\begin{tabular}{l c ccccccccccc}
		\toprule
		& Source & \multicolumn{11}{c}{Target} \\ \cmidrule(lr){2-2} \cmidrule(lr){3-13}
		& \rotbox{ImageNet} & \rotbox{Caltech101} & \rotbox{OxfordPets} & \rotbox{StanfordCars} & \rotbox{Flowers102} & \rotbox{Food101} & \rotbox{Aircraft} & \rotbox{SUN397} & \rotbox{DTD} & \rotbox{EuroSAT} & \rotbox{UCF101} & \rotbox{Average} \\
		\midrule

		CoCoOp   &70.13 &93.33	&88.76	&64.49	&69.00	&85.48	&19.09	&64.03	&42.58	&45.61	&66.43	&63.88\\
		\hline
		CoOp  &69.86 &93.70	&89.14	&64.51	&68.71	&85.30	&18.47	&64.15	&41.92	&45.39	&66.55	&63.78\\
		\textbf{+ GRAM} &\textbf{70.49}		&\textbf{\color{WildStrawberry}{94.98}}	&\textbf{90.94}	&\textbf{64.80}	&\textbf{69.08}	&\textbf{85.62}	&\textbf{19.60}	&\textbf{64.17}	&\textbf{41.33}	&\textbf{46.56}	&\textbf{66.66}	&\textbf{64.37}\\
		
		\hline
		VPT &70.11	&93.30	&87.76	&62.64	&67.91	&83.64	&20.62	&64.51	&41.21	&40.85	&63.37	&62.58\\
		
		\textbf{+ GRAM} &\textbf{70.46}		&\textbf{93.94}	&\textbf{88.04}	&\textbf{63.42}	&\textbf{67.86}	&\textbf{83.81}	&\textbf{21.27}	&\textbf{64.47}	&\textbf{41.67}	&\textbf{41.51}	&\textbf{63.54}	&\textbf{62.95}\\
		\hline
		\textbf{UNIGRAM} &\color{WildStrawberry}{70.84} &93.69 &\color{WildStrawberry}{91.37}	&\color{WildStrawberry}{64.84}	&\color{WildStrawberry}{69.54}	&\color{WildStrawberry}{85.99}	&\color{WildStrawberry}{21.56}	&\color{WildStrawberry}{64.69}	&\color{WildStrawberry}{42.60}	&\color{WildStrawberry}{46.61}	&\color{WildStrawberry}{66.68}	&\color{WildStrawberry}{64.76}\\
		\bottomrule
	\end{tabular}
\end{table*}

\subsection{Generalization From Base to New Classes}~\label{base-to-new}
Prompt tuning with a few training samples (16 shots) of the base classes, we evaluate the adaptation ability of models on the remaining testing samples of the base classes and the generalization ability of models on the unseen classes. Table~\ref{t1} summarizes the results. \textbf{(1)} Overall, the proposed GRAM method is capable of generalizing to different baseline models, ranging from textual prompt tuning to visual prompt tuning. Our GRAM can not only boost their few-shot adaptation ability on the base classes but also significantly improves their generalizability on the unseen classes. \textbf{(2)} As for the average accuracy of the base classes over 11 datasets, our GRAM largely improves CoOp and VPT  by 1.16\% and 1.51\%, respectively, indicating that our unsupervised meta-learning empowers the adaptation ability of existing methods. \textbf{(3)} As for the average accuracy of the new classes over 11 datasets, our GRAM brings about 1.82\% and 1.87\% improvements on CoOp and VPT, respectively, which demonstrates that the proposed gradient regulating function can effectively mitigate the overfitting problem. \textbf{(4)} Moreover, our GRAM enables the visual and textual prompt tuning to work in a mutually-enhanced way. Our UNIGRAM achieves stronger few-shot generalization performance beyond its uni-modal components, improving the average accuracy of unseen classes from 73.11\%~(CoOp) to 75.92\%. Note that, UNIGRAM largely outperforms CoCoOp by 14.29\%, 6.38\%, and 11.34\% on FGVCAircraft, DTD, and  EuroSAT datasets, respectively.

\vspace{-0.05cm}
\subsection{Cross-Domain Generalization}~\label{cross-domain}
\vspace{-0.35cm}

Contrastively pre-trained vision-language models have demonstrated strong generalizability, but prompt tuning on limited data from a specific dataset might undermine the generalizability of the pre-training models. In this section, we evaluate the out-of-distribution generalization performance of prompt tuning methods. Following~\cite{zhou2022conditional}, we evaluate the cross-domain generalization performance by transferring the prompts learned from ImageNet to four other variants of ImageNet with domains shift.

As shown in Table~\ref{t2}, our GRAM consistently improves the domain generalizability of CoOp and VPT on all target datasets while at the same time maintaining a higher performance on the source dataset. This indicates that meta-learning to regulate the gradient can effectively prevent the models from overfitting to some spurious correlations of a single domain. In addition, GRAM brings about clear improvement by harmonically combining visual and textual prompts. On the ImageNet-R, UNIGRAM significantly surpasses CoCoOp by 1.16\%.

\subsection{Cross-Dataset Generalization}~\label{cross-dataset}
We further consider a more challenging setting, that is, generalizing across different datasets. The models are only prompt tuned on the source dataset and required to transfer to other 10 datasets in a zero-shot manner. As illustrated in Table~\ref{t3}, equipped with our GRAM, the average transfer performance of CoOp and VPT increases 1.20\% and 1.08\%, respectively. This validates that GRAM can also improve the cross-dataset generalizability of different methods. Further, our UNIGRAM not only achieves the highest performance on the source dataset but also demonstrates stronger cross-dataset generalization performance over existing methods, outperforming CoCoOp by 1.97 points.

\subsection{Extremely Few-Shot Generalization}
We further consider extremely few-shot scenarios to better evaluate the adaptation and generalization abilities of our approach. We measure the 4-shot performance instead of the 16-shot performance, where we keep the same training details and evaluation metrics as the 16-shot setting. We report the cross-domain and cross-dataset generalization performance in Table~\ref{ap_t1} and Table~\ref{ap_t2}. When the training samples are extremely limited, we find that our GRAM also demonstrates a strong ability to boost the cross-domain and cross-dataset generalizability of  CoOp and VPT while maintaining a higher performance on the source dataset.  Furthermore, our UNIGRAM exhibits superior generalizability over existing methods by harmonically combining the visual and textual prompt tuning.

\subsection{In-Depth Analysis}
\begin{table}[!t]
	\centering
	\caption{Ablation results (\%) over 11 datasets.}
	\vspace{-0.3cm}
	\label{t4}
	\resizebox{\linewidth}{!}{
	\begin{tabular}{ ll ccc}
		\toprule
		& & Base & New &H\\
		\hline
		0&\textbf{UNIGRAM}  &80.34  &75.92   &78.07  \\
		1&\quad-domain shift simulating  &79.68  &75.56  &77.57 \\
		2&\quad-gradient regulating &77.90  &74.79  &76.31\\
		3&\quad-meta-learning = prompt pre-training &75.90  &74.32  &75.10\\
		4&\quad-CHC = supervised meta-learning&78.68  &74.47  &76.52\\
		\hline
		5& \textbf{joint textual\&visual prompt tuning}  &77.41    &72.85   &75.06    \\
		\bottomrule
	\end{tabular}
}
\vspace{-0.3cm}
\end{table}

\noindent
\textbf{Effectiveness of Individual Components.} In Table~\ref{t4}, we train the following ablation models: (1)~w/o domain shift simulating: support set samples are uniformly sampled from all domains, without simulating domain shift. (2)~w/o gradient regulating: we remove the gradient regulating function and update the model using raw gradient. (3)~w/o meta-learning: we remove the bi-level meta-learning paradigm and directly use image classification over clustering data as a ``pre-training'' task to learn a better soft prompt initialization. (4)~w/o Cross-Modal Hierarchical Clustering~(CHC): instead of using unlabeled data, we directly use an annotated image classification dataset~(\ie, WebVision~\cite{li2017webvision}) for meta-training. We construct meta-training tasks by sub-sampling from the set of classes. (5)~joint textual\&visual prompt tuning: we further consider a straightforward approach that tunes the visual and textual prompts jointly. 

The results of Row~1 indicate that using our proposed cross-modal hierarchical clustering to simulate domain shift is crucial for our gradient regulating function to learn to avoid overfitting. Without domain shift simulating, the generalization performance on the new classes is degraded. Also, Row~2 validates the superiority of the proposed gradient regulating function on preventing the model from being misled by some domain-specific correlations. Our gradient regulating function takes up 15\% of the relative gain on accuracy (new classes). Further, the results of Row~3 show that the main performance gain does not directly come from the pre-training image-text pairs. Instead, our GRAM provides a novel way to utilize the unlabeled data to address the limitations of prompt tuning. Then, according to Row~4, we notice that the large-scale unlabeled data is a better choice for meta-learning, which covers a wider range of semantics and domains than existing supervised datasets. Finally, from the results of Row~5, we observe that direct joint tuning of the textual and visual prompts performs slightly worse than CoOp. In contrast, Row~0 demonstrates that our approach enables the visual and textual prompt tuning to work in a mutually-enhanced way.

\begin{figure}[!t]
	\centering
	\includegraphics[width=\linewidth]{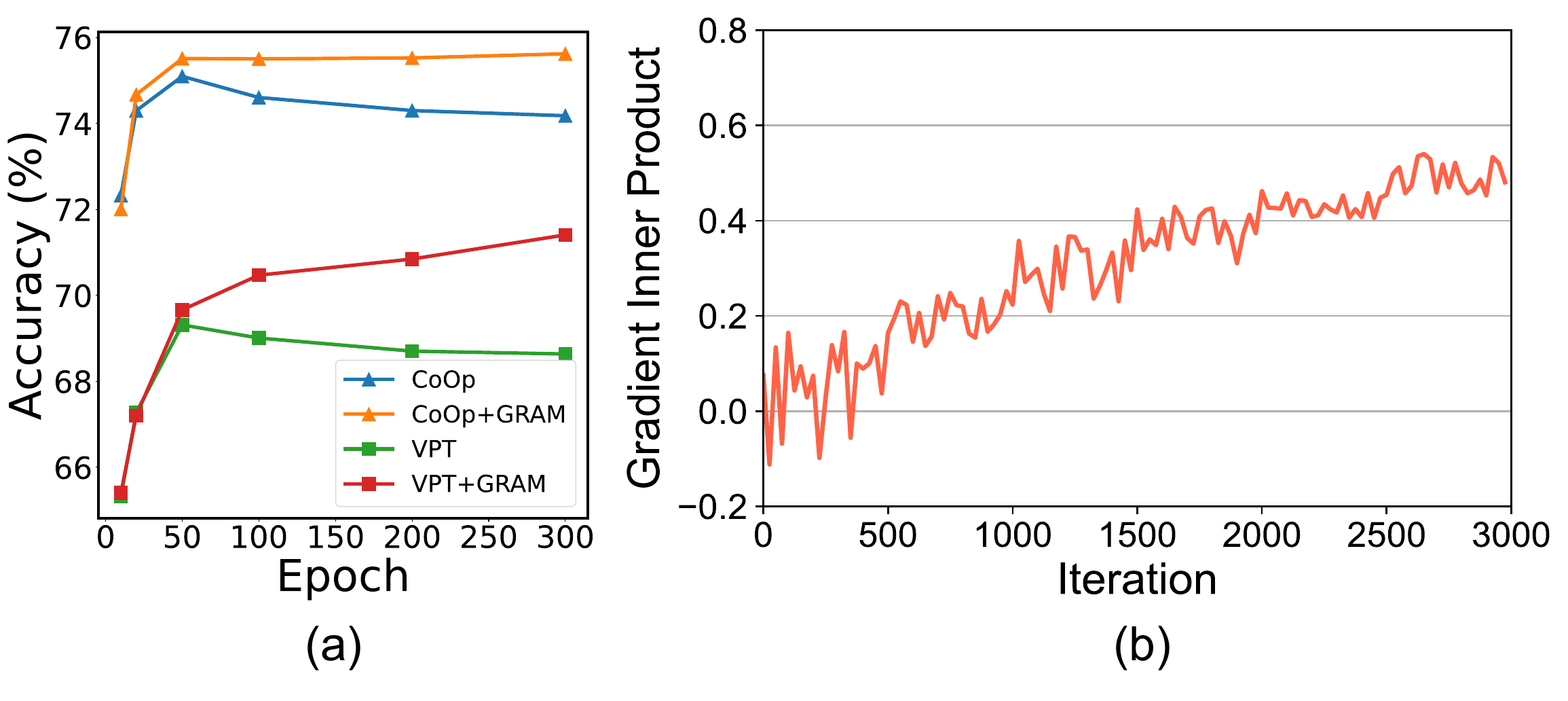}
	\vspace{-0.4cm}
	\caption{(a) Test accuracy during training. (b) Normalized gradient inner product during training.}
	\label{ablation}
	\vspace{-0.4cm}
\end{figure}

\noindent
\textbf{Analysis on Adaptation and Generalization.} We report the averaged few-shot performance per epoch. As shown in Figure~\ref{ablation}(a), CoOp+GRAM and VPT+GRAM can better adapt to testing datasets, demonstrating the stronger adaptation ability brought by GRAM. Besides, as the training continues, the performance of CoOp and VPT on the new classes drops seriously. In contrast, the proposed gradient regulating function effectively prevents CoOp+GRAM and VPT+GRAM from overfitting to training data.

\begin{table}[!t]
	\centering
	\vspace{-0.2cm}
	\caption{Ablation results (\%) with respect to different prompt token numbers over 11 datasets.}
	\label{ap_t3}
	\begin{tabular}{c ccc}
		\toprule
		Prompt Token Number& Base & New &H\\
		\hline
		2  &77.49  &76.09  &76.78 \\
		4  &80.34  &75.92  &78.07\\
		6  &80.09  &75.63  &77.80\\
		8  &80.31  &74.79  &77.45\\
		\bottomrule
	\end{tabular}
\end{table}

\noindent
\textbf{Visualization of Gradient Regulating.} To verify whether our gradient regulating function can regulate the gradient conflict between support set and query set, we report the normalized gradient inner product between support set and query set during training. As shown in Figure~\ref{ablation}(b), we clearly observe that the normalized gradient inner product gradually increases during training. This indicates that our gradient regulating function is learned to regulate the gradient over the support set into a more generalizable direction.

\noindent
\textbf{Analysis on the Number of Prompt Tokens.} We explore the impact of different numbers of the learnable prompt tokens by varing the number of prompt tokens from 2 to 8. We report the average accuracy over 11 datasets in Table~\ref{ap_t3}. By increasing the number of prompt tokens from 2 to 4, the performance on the base classes is clearly improved while the performance on the new classes drops slightly. Then, continuing increasing the token numbers will damage the generalization performance on the new classes.

\section{Conclusions}
In this paper, we point out the initialization-sensitive issue and the generalizability degradation issue of current prompt tuning methods for few-shot generalization. We introduce a model-agnostic meta-prompting method GRAM, which jointly learns an efficient soft prompt initialization for better adaptation and a lightweight gradient regulating function for strong cross-domain generalizability using only unlabeled image-text pairs. Extensive experiments on several settings (\eg, cross-domain generalization, cross-dataset generalization) over 11 datasets demonstrate that GRAM can boost existing methods in a plug-and-play fashion. Further experiments show that our GRAM enables both visual and textual prompts to work in a complementary way, exhibiting stronger few-shot generalization ability.

\section*{Acknowledgment}
This work has been supported in part by the National Key R\&D Program of China (2022ZD0160101), Zhejiang NSF (LR21F020004),  the NSFC (No. 62272411),  Alibaba-Zhejiang University Joint Research Institute of Frontier Technologies, and Ant Group. We thank all the reviewers for their valuable suggestions.

\clearpage
{\small
\bibliographystyle{ieee_fullname}
\bibliography{egbib}
}

\end{document}